\documentclass[conference]{IEEEtran}
\IEEEoverridecommandlockouts
\usepackage[utf8]{inputenc}
\usepackage{dirtytalk}
\usepackage{flushend}
\def\BibTeX{{\rm B\kern-.05em{\sc i\kern-.025em b}\kern-.08em
    T\kern-.1667em\lower.7ex\hbox{E}\kern-.125emX}}

\begin{document}
\bstctlcite{IEEEexample:BSTcontrol}

\title{The case for psychometric artificial general intelligence}

\author{\IEEEauthorblockN{Mark McPherson}
\IEEEauthorblockA{\textit{Dept. of Computing and Informatics} \\
\textit{Bournemouth University} \\
Bournemouth, UK \\
https://orcid.org/0000-0003-1932-2344}}

\maketitle

\begin{abstract}
A short review of the literature on measurement and detection of artificial general intelligence is made. Proposed benchmarks and tests for artificial general intelligence are critically evaluated against multiple criteria. Based on the findings, the most promising approaches are identified and some useful directions for future work are proposed.
\end{abstract}

\begin{IEEEkeywords}
artificial general intelligence, AGI, psychometric AI, PAI, PAGI
\end{IEEEkeywords}

\section{Introduction}
The roots of artificial intelligence (AI) can be found in the Dartmouth Summer Research Project of 1956 \cite{b1}, organised with the hope of making a significant advance in programming intelligence into a computer over the course of 2 months. While the problem the participants set out to solve has been found to be vastly more complex than they imagined, in the decades since that summer project the field has made huge advances. AIs can now perform at expert human level, or beyond, in several tasks. Some of the most well-known include the game of Go \cite{b2}, modern computer games such as StarCraft \cite{b3}, image classification \cite{b4}, language translation \cite{b5} and medical diagnosis \cite{b6}. We may well be on the cusp of an era in which AIs will prove to be better at more tasks than humans. But, if one of these superhuman AIs were to be placed in an environment only slightly different to the one it usually operates in, the strong likelihood is that it would cease to perform its task effectively, if at all \cite{b7}. Despite the great deal of skill current AIs have attained, they cannot match the human, or even animal, ability to adapt to almost any environment they find themselves in. Humans can often see a new task performed only once or twice and proceed to be able to repeat the task. In fact, often enough they can perform a task, on the first attempt, without even seeing it performed first (e.g. someone encountering a new device does not always need the instructions or a demonstration to use it correctly). This is the case for an inconceivable number of tasks. No AI we have today displays such adaptability.
There are several classes in which AI is usually placed. The most used terms, introduced by Searle \cite{b8}, are ‘strong’ AI and ‘weak’ AI. The distinction between strong and weak is mainly a philosophical one, concerned with whether an AI is capable of understanding in the same way a human can, or merely act like it does, respectively. The terms ‘narrow’ AI and ‘general’ AI are also used. These classifications refer to the scope of ability of an AI. A ‘narrow’ AI is only capable of performing tasks within a small number of domains (usually only a single domain). This is the level of AI we can currently create. AlphaGo, for example, has only been shown so far to perform extremely well in the domain of board games. ‘General’ AI is an AI that can perform tasks across a huge number of domains, even domains which were unknown at the time of its creation – it displays general intelligence. The closest examples we have to general intelligence are not artificial agents, but ourselves – humans demonstrate general intelligence. An AI demonstrating the same is termed an artificial general intelligence (AGI).
In order to compare the level of general intelligence between agents, we must use some sort of benchmark. To paraphrase the eminent physicist Lord Kelvin \cite{b9} \textit{\say{When you can measure a thing, you know something about it; when you cannot measure it, your knowledge is of an unsatisfactory kind}}.

\section{Test Proposals}

Several surveys of proposed definitions of AGI, along with tests, benchmarks and frameworks for detecting and measuring it already exist \cite{b10,b11,b12}. Here we cover some newer and some older but not widely surveyed ideas.

Mikhaylovskiy \cite{b13} propose 6 tests. Their Explanation Test requires an AI to provide an explanation and quantitative characteristics of an empirical phenomenon, given a well-defined scientific theory concerned with said phenomenon. Next, their Problem-Setting task requires an AI to be able to produce a problem which could be given in the first task. The Refutation test asks for an AI to be able to devise an experimental methodology that could determine the best model of some phenomena from a set of competing models. The New Phenomenon Prediction test tasks an AI with producing a new prediction, given some well-defined scientific theory. Their Business Creation test looks for an AI being capable of creating a successful business start-up. Finally, to pass their Theory Creation test, an AI would be required to produce a new and better scientific theory in some research field.

Chollet \cite{b14} proposes a benchmark dataset, influenced largely by the field of psychometrics, named the Abstraction and Reasoning Corpus (ARC). Chollet argues that any approach to AGI should be human-centric. Accordingly, the ARC benchmark was created with the over-arching goal of providing a method of comparing any two agents, human or artificial, in a fair way that quantifies what he calls their ‘generalisation power’. By fair, ARC stipulates the prior knowledge that is assumed of an agent. The prior knowledge is a set of core knowledge priors, innate to humans (or acquired very early in life) \cite{b15}.

Potapov et al \cite{b16} propose as a benchmark the Primary School Olympiad Math Tasks (PSOMT), used in America as an educational assessment for young children.

For Bringsjord et al. \cite{b17} the crux of intelligence is assumed to be in the ability to be creative. This was the view espoused by Lady Lovelace, and the proposed test is named after her. To pass the Lovelace Test, an artificial agent must be able to consistently ‘surprise’ it’s creators, in the sense that the artifacts the agent produces are not explainable.

A test posited by Bringsjord and Licato \cite{b18} consists of a hypothetical room (termed the Piaget-MacGyver Room (PMR), after some of the thinkers that influenced its conception) containing a fixed set of ingredients. An agent, assumed to know what the set of ingredients is before being tested, is said to exhibit general intelligence if, and only if, it can pass any test constructed of the ingredients.

Several ‘practical’ tests are outlined by Goertzel et al \cite{b19}. The common theme here is that each test involves completion of some common human task, such as making a coffee or understanding a story. These tasks presumably require intelligence for a human to perform successfully, so as per Minsky’s definition \cite{b20} of AI successfully completing such a task would indicate intelligence, but not necessarily general intelligence.

Nilsson’s Employment Test \cite{b21} proposes the measuring progress towards human-level machine intelligence (HLMI), rather than general intelligence (hence side-stepping the issue of just what general intelligence is). The metric here is the percentage of all jobs in which people may be employed which can be successfully performed by artificial agents.

\section{Critical Evaluation}
There are many factors that are important to consider when evaluating how well some test, framework or benchmark can drive progress in AGI. We will use factors previously considered by Legg and Hutter \cite{b11}.
\begin{itemize}
\item Validity
\item Informativeness
\item Range
\item Generality
\item Dynamism
\item Bias
\item Fundamentalism
\item Formality
\item Objectiveness
\item Completeness of definition
\item Universality
\item Practicality
\end{itemize}

In broad terms, the proposals in II can be split into two classes – tests and benchmarks.

\subsection{Tests}\label{TESTS}
All but ARC and PSOMT are tests. A specific task is given and an agent either passes the test by performing the task successfully, or it fails to pass the test.

\subsubsection*{Validity}
A prerequisite of general intelligence is being able to adapt to unforeseen environments. Since the ability of an agent to perform a single task is being tested, there is no way to determine if general intelligence is being displayed. 
\subsubsection*{Informativeness}
There is no scale of measurement – an agent either passes or it does not. Although gradual improvement can be seen qualitatively, such tests are not informative in how well an agent does. Comparing multiple agents is difficult, even impossible for those with a similar level of ability.
\subsubsection*{Range}
Since these tests provide no quantitative measure, the question of range of intelligence is moot. However, all such tests involve a task that is one typical of humans.
\subsubsection*{Generality}
While it may be possible to test any agent, whether biological or artificial, doing so for an agent belonging to the lower animal kingdom or an artificial agent not purported to be an AGI by its creators would reveal nothing useful.
\subsubsection*{Dynamism}
It is not immediately obvious that such tests could offer some degree of dynamism. There is a possibility of cycling between testing and feeding the result back to an agent. But since the feedback is of a binary nature, it is doubtful to be of any use – the agent would be attempting to get better at the test based on the feedback that it had either failed, a poor signal in the circumstances, or that it had actually already passed!
\subsubsection*{Bias}
Such tests are inevitably biased towards a human-like intelligence, since the tasks they ask an agent to perform are taken from the set of tasks we think of, from our anthropocentric perspective, as requiring intelligence.
\subsubsection*{Fundamentalism}
For each test, an assumption is made that the task they set requires general intelligence. These assumptions are arguably, in many cases, too strong. It is possible that our understanding of general intelligence may be refined in future. Some tasks generally seen as requiring general intelligence today may be ‘downgraded’ in future. So, these tests are not fundamental.
\subsubsection*{Formality}
None of these tests are expressed precisely, in formal mathematical language. As such they are open to interpretation.
\subsubsection*{Objectiveness}
Whether an agent passes or not is decided by one or more human judges. This makes the tests subjective.
\subsubsection*{Completeness of definition}
None of the tests are fully defined, in the sense of being unambiguous and detailing exactly the test environment and ‘rules’ of the test.
\subsubsection*{Universality}	As already described, these tests are human-centric due to the selection of tasks from those tasks of interest to humans.
\subsubsection*{Practicality}
There are several factors preventing these tests from being practicable ones. Due to this unpractical nature, such tests cannot be used for driving progress towards AGI. They rely on some element of human oversight; hence they cannot be automated. In addition, this human oversight introduces subjectivity and hampers the replicability of the tests. Implementing the tests would be a costly endeavour owing to the need to pay the judges and the speed would be many orders of magnitude slower than a fully automated framework.

\subsection{Benchmarks}
ARC and PSOMT fall into the area of research known as psychometric artificial general intelligence (PAGI). Psychometric testing is a well-studied and mature field within psychology. PAGI tests agents by using a battery of tasks, often within a range of domains.

\subsubsection*{Validity}
The results of testing an agent can be shown statistically to be correlated to performance in future tests. One feature present in some PAGI approaches is that the precise tasks in the test are not known to an agent (or its creators, if they exist) in advance. Thus, the ability of the agent to perform well at a previously unseen task is tested.
\subsubsection*{Informativeness}
The result of testing using a PAGI approach is an actual number. This means that two agents can be directly compared in a quantitative manner. Also, small increments in ability can be easily seen; this is important for driving progress.
\subsubsection*{Range}
A wide range of ability levels can be measured. How wide depends on the exact composition of the benchmark, such as the number of distinct tasks. The range of applicability goes from a level of ability capable of success at just one task up to a maximum level of ability capable of success at all tasks. So, an agent unable to perform any of the possible tasks, or all of them, would not be a suitable candidate for a benchmark.
\subsubsection*{Generality}
Unlike the task-specific case, we can sensibly submit any type of agent to such benchmarks. As noted for range, we would need to be careful in selecting the specific tasks comprising the benchmark, to ensure they are applicable to the agent.
\subsubsection*{Dynamism}
When the tasks that comprise a benchmark are not known in advance, the benchmark can test the ability of an agent to learn and adapt. This allows an agent to formulate and try approaches to a task, rather than being a snapshot of ‘crystallised’ ability.
\subsubsection*{Bias}
It is possible to formulate a benchmark wherein the tasks do not depend on abilities or knowledge specific to one type of agent. In addition, any prior knowledge or experience assumed can be explicitly stated. To be meaningful, an agent taking the test must satisfy these assumptions.
\subsubsection*{Fundamentality}
Just as in traditional human, psychometric testing, results can be normalized. So, although our understanding of what constitutes general intelligence, and to what degree a factor or ability is important to it, the results of a benchmark can be consistent in a statistical sense. Benchmarks can thus evolve naturally over time, just as our understanding does.
\subsubsection*{Formality}
Since we wish to avoid testing anything that lends bias, prime candidates for the type of tasks to use are mathematical. The tasks themselves and the score an agent achieves can be simply defined in formal mathematics.
\subsubsection*{Objectiveness}
With a formally defined method of scoring an agent, all need for subjective judgement is removed. This ensures replicability.
\subsubsection*{Completeness of definition}
A PAGI benchmark can be specified in explicit and unambiguous formal mathematics. Thus, any benchmark has an easily formed and complete definition.
\subsubsection*{Universality}
Although it is likely that the most useful benchmarks will involve tasks that are of interest to humans, they do not need to be. Benchmark creators are free to develop any tasks and the method by which they are scored.
\subsubsection*{Practicality}
PAGI inspired benchmarking is extremely conducive to practical use. A benchmark requires no human supervision. Being formally defined and objective it may easily be encoded into machine readable form and the results can be presented in a standardized manner for easy comparison against other agents.

\section{Conclusion}
The evaluation in the previous section has shown how a psychometric approach to AGI measurement surpasses task-specific approaches in every consideration. While there is no claim that it is the perfect framework to use, it can provide a guiding light towards faster progress in the development of ever more general artificial agents. Attempting to use a task-specific approach is akin to skipping learning how to walk, in the hope that learning to run will teach the simpler skill.

PAGI inspired benchmarks could be incorporated into an online repository. With some standard interface, such a repository could allow researchers and developers to, in real-time or close to real-time, obtain results for the systems they are engaged with across a range of benchmarks. We expect the existence of such a repository would result in a noticeable increase in momentum in the field of AGI, with knock-on effects to the, at present more useful (economically speaking), predominant field of narrow AI and its applications.

\bibliographystyle{IEEEtran}
\bibliography{case_for_PAGI}

\end{document}